\documentclass[letterpaper, 10 pt, conference]{ieeeconf}  

\IEEEoverridecommandlockouts                              

\usepackage{amsmath} 
\usepackage{amssymb}  
\usepackage{xcolor}
\usepackage{graphicx}
\usepackage{subfig}

\title{\LARGE \bf
Distributed Heuristic Multi-Agent Path Finding with Communication}

\author{Ziyuan Ma$^{1*}$, Yudong Luo$^{1*}$ and Hang Ma$^{1}$
\thanks{$^{1}$School of Computing Science,
        Simon Fraser University, Canada
        {\tt\small \{ziyuan\_ma, yudong\_luo, hangma\}@sfu.ca}}%
\thanks{* indicates equal contribution.}
}

\begin{document}

\maketitle
\thispagestyle{empty}
\pagestyle{empty}

\begin{abstract}

Multi-Agent Path Finding (MAPF) is essential to large-scale robotic systems. Recent methods have applied reinforcement learning (RL) to learn decentralized polices in partially observable environments. A fundamental challenge of obtaining collision-free policy is that agents need to learn cooperation to handle congested situations. This paper combines communication with deep Q-learning to provide a novel learning based method for MAPF, where agents achieve cooperation via graph convolution. To guide RL algorithm on long-horizon goal-oriented tasks, we embed the potential choices of shortest paths from single source as heuristic guidance instead of using a specific path as in most existing works. Our method treats each agent independently and trains the model from a single agent's perspective. The final trained policy is applied to each agent for decentralized execution. The whole system is distributed during training and is trained under a curriculum learning strategy. Empirical evaluation in obstacle-rich environment indicates the high success rate with low average step of our method.

\end{abstract}

\section{INTRODUCTION}
Multi-Agent Path Finding~\cite{stern2019multi} is a path arrangement problem for a team of agents. Each agent is required to move from its given start location to its given goal location while avoiding collisions with other agents. MAPF arises in many real world applications of multi-agent systems, such as warehouse and office robots~\cite{wurman2008coordinating,veloso2015cobots}, aircraft-towing vehicles~\cite{morris2016planning}, and video games~\cite{silver2005cooperative,ma2017feasibility}.

MAPF is NP-hard to solve optimally on graphs~\cite{yu2013structure} and even 2D grids~\cite{banfi2017intracatability}. Traditional centralized planning methods reduce MAPF to other well-studied problems, e.g., ILP~\cite{yu2013planning} and SAT~\cite{surynek2016efficient}, or solve it with search-based algorithms, e.g., enhanced A* search in joint state space~\cite{goldenberg2014enhanced,wagner2015subdimensional}, Conflict-Based Search~\cite{sharon2015conflict} and their improved variants~\cite{li2019improved}. The limitation of these centralized planning methods is that they do not scale well to a large number of agents.

Recently, RL with decentralized execution has been applied to address this issue. During execution, each agent follows a policy individually and makes decisions based on its local observations. A common approach is to train a reactive policy that corrects actions of an agent to avoid collisions during execution~\cite{chen2017decentralized,long2018towards}. However, such policies often lead to deadlocks or livelocks in cluttered and tight environments~\cite{sartoretti2019primal}. State-of-the-art methods guide RL with imitation learning (IL)~\cite{sartoretti2019primal} or a single-agent shortest path~\cite{liu2020mapper}. Nevertheless, the environment settings and the algorithmic designs of these methods still need to be improved to allow for using RL effectively in the standard MAPF setting~\cite{stern2019multi} for cluttered and tight environments. For example, \cite{sartoretti2019primal} assumes that agents take actions one at a time, and \cite{liu2020mapper} assumes that agents are removed from the environment upon reaching their goal locations. These assumptions differ from the standard MAPF setting and simplify the problem. The resulting policies are not trained to handle collisions when agents move simultaneously or to handle deadlocks and livelocks when an agent reaches its goal location but obstructs other agents from getting to their goal locations. Moreover, these methods can also be improved to use guidance more intelligently. For instance, \cite{sartoretti2019primal} generates expert demonstrations with a centralized MAPF planner, which requires solving a MAPF problem instance and is thus time-consuming. Also, \cite{liu2020mapper} gives extra shaped rewards to each agent that incentivize the agent to follow one single-agent shortest path, which can mislead the agents since paths to their goal locations are not unique and forcing the agents to follow paths individually could harm their cooperation in multi-agent settings.

To tackle the above issues, in this work, we propose a deep Q-network with single-agent heuristic guidance and multi-agent communication to facilitate long-horizon path finding and cooperation of agents. We follow the standard MAPF setting~\cite{stern2019multi} where agents move simultaneously and are kept in the environment after reaching their goals. Instead of providing a specific path as guidance and modifying the individual rewards accordingly, we embed all the potential choices of shortest paths from single source as the heuristic guidance in the input to the Q-network, and the model learns rational knowledge from the heuristic by itself. 
As it is also crucially important for agents to learn cooperation, we formalize the multi-agent environment as a graph and let agents communicate with neighbors via graph convolution. Multi-head attention is employed as convolutional kernel to extract the relation representation between agents. We treat each agent independently and leverage single-agent Q-learning for multi-agent partially observable Markov game without learning a joint action value, making it easy to scale. To speed up learning and enhance the performance, our system is distributed during training based on the Ape-X framework~\cite{horgan2018distributed} and is trained under a curriculum learning~\cite{bengio2009curriculum} strategy. During decentralized execution, each agent adopts the same policy and moves simultaneously. Empirical results show the high success rate with low average step of our method compared with its counterpart.

\textbf{Contributions}. Our main contributions are summarized as follows.
\begin{enumerate}
\item A learning environment close to standard MAPF setting, where agents move simultaneously and are kept in the environment upon reaching the goals. 
\item A novel heuristic deep Q-learning method with graph convolutional communication for goal-oriented path finding and cooperation of agents.
\item A distributed independent Q-learning method for partially observable Markov games, scaling to large amount of agents.
\end{enumerate}

\section{PROBLEM DEFINITION}
The MAPF problem is formalized as follows. Given an undirected graph $G=(V,E)$ and an agent set $N$, each agent $i$ is assigned a unique start vertex $s_i\in V$ and a unique goal vertex $g_i \in V$. At each discrete time step $t = 0,\ldots ,\infty$, each agent can either move to an adjacent vertex or wait at its current vertex. A path for agent $i$ contains a sequence of adjacent (indicating a moving) or identical (indicating a waiting) vertices beginning at the start vertex $s_i$ and terminating at the goal vertex $g_i$. A collision between agents is either a vertex collision, which is a tuple $\langle i, j, v, t\rangle $ where agents $i$ and $j$ reaching at the same vertex $v$ at time $t$, or an edge collision, defined as a tuple $\langle i, j, u, v, t\rangle $ where agents $i$ and $j$ traverse the same edge $(u,v)$ in opposite directions at time $t$. A solution to MAPF is a set of collision-free path, one for each agent. The quality of a solution is measured by the sum of arrival time of all agents at their goal vertices. We focus on 2D 4-neighbor grids, while our method can be easily generalized to higher dimensions.


\section{BACKGROUND}
\subsection{Partially Observable Markov Game}
We consider the MAPF problem as a Markov game~\cite{littman1994markov} with partial observability, represented as a tuple $\langle N,\mathcal{S},\{\mathcal{A}_i\},\{\mathcal{O}_i\},\{\mathcal{R}_i\},\mathcal{P},\gamma\rangle $. $N$ is a finite set of agents, indexed by $1,\ldots ,n$. $\mathcal{S}$ is a finite state set. $\mathcal{A}_i$ is a finite action set available to agent $i$, and $\mathcal{A}=\mathcal{A}_1 \times \ldots \times \mathcal{A}_n$ is the set of joint actions. $\mathcal{O}_i$ is a finite observation set for agent $i$, and $\mathcal{O}=\mathcal{O}_1\times \ldots \times \mathcal{O}_n$ is the set of joint observations. $\mathcal{P}(s',\vec{o}|s,\vec{a})$ is the state transition and observation probability function, where $\vec{a}$ and $\vec{o}$ are instances of $\mathcal{A}$ and $\mathcal{O}$. $\mathcal{R}_i:\mathcal{S}\times\mathcal{A}\rightarrow\mathbb{R}$ is the reward function for agent $i$. $\gamma$ is the discount factor. Agents simultaneously choose an action and receive a reward and an observation from the environment. The aim for each agent is to maximize its expected total return during the game. 

\subsection{Q-Learning and Deep Q-Networks (DQN)}
Q-Learning and DQN~\cite{mnih2015human} are popular methods in single-agent and fully-observable RL settings. An agent observes current state $s_t\in\mathcal{S}$ and selects an action $a_t\in\mathcal{A}$ according to a policy $\pi$ at each time step $t$. The agent's objective is to maximize the expectation of discounted total return $R_t = r_t+\gamma r_{t+1}+\gamma^2r_{t+2}+\ldots $, where $r_t$ is the reward received at time $t$. Q-Learning utilizes an action value function for policy $\pi$ as $Q^{\pi}(s,a)=\mathbb{E}[R_t|s_t=s, a_t =a]$ and can be recursively defined by $Q^{\pi}(s,a)=\mathbb{E}_{s'}[r+\gamma\mathbb{E}_{a'\sim\pi}[Q^{\pi}(s',a')]]$. The optimal action value, $Q^*(s,a)=\max_{\pi}Q^{\pi}(s,a)$, satisfies the Bellman optimality equation $Q^*(s,a)=\mathbb{E}_{s'}[r+\gamma\max_{a'}Q^*(s',a')|s,a]$. DQN learns the action value function using neural networks parameterised by $\theta$, represented as $Q(s,a;\theta)$. The optimal policy is trained by minimizing the loss  $\mathcal{L}(\theta)=\mathbb{E}_{s,a,r,s'}[(Q^*(s,a;\theta)-y)^2]$, where $y=r+\gamma\max_{a'}Q^*(s',a';\bar{\theta})$. Here, $\bar{\theta}$ is the parameter of a target network and is updated periodically with the most recent $\theta$. In partially observable environment, agents in general need to condition on an observation-action history. In this setting, DQN is equipped with a recurrent unit such as LSTM~\cite{hochreiter1997long} or GRU~\cite{chung2014empirical}.

\subsection{Independent Q-Learning (IQL)}
Q-Learning can be directly extended to multi-agent settings by each agent learning its own action value function $Q^i$ independently and simultaneously, treating other agents as part of the environment~\cite{tan1993multi}. Empirical results have shown that IQL works well in some multi-agent settings like two-player pong~\cite{tampuu2017multiagent}.

\section{RELATED WORK}

\subsection{Multi-Agent Reinforcement Learning (MARL)}
Apart from IQL, the framework of centralized training with decentralized execution has been widely adopted by MARL. These approaches can be divided into two categories, value-based and actor-critic methods. For value-based methods, value decomposition networks (VDN)~\cite{sunehag2018value} proposes to have separate action value function for multiple agents when only one shared team reward is available. It aims to learn a joint action-value function by summing up individual action-value linearly. QMIX~\cite{rashid2018qmix} extends VDN by mixing individual action values in a nonlinear way. For actor-critic methods, counterfactual multi-agent (COMA) policy gradient~\cite{foerster2018counterfactual} learns a single centralized critic for all agents to estimate Q-function and multiple decentralized actors to optimize policies. In order to address multi-agent credit assignment, it estimates a counterfactual advantage value for each agent. In contrast with COMA, Multi-Agent DDPG (MADDPG)~\cite{lowe2017multi} learns separate actors and critics for different agents. The centralized critic of each agent has access to observations and actions of all the agents during learning. It should be noted that the above methods are all applied to fully cooperative Markov game, where agents share the same reward function, different from our setting, where agents have their individual ones.

\subsection{Reinforcement Learning based MAPF}

Deep RL  has achieved great success in single-agent path planning~\cite{tamar2016value}. Recently, solving MAPF via RL approach has drawn great attention to researchers. A well known framework is called PRIMAL~\cite{sartoretti2019primal}, which is a hybrid method combining RL and IL to learn a decentralized policy in partially observable environment. The RL part relies on the asynchronous advantage actor critic (A3C) network~\cite{mnih2016asynchronous}, where each agent (thread) shares the same global parameters. The IL part is simply behaviour cloning, requiring training data generated from a centralized planner. The idea of utilizing centralized planner as guidance is also adopted by Global-to-Local Autonomy Synthesis (GLAS)~\cite{riviere2020glas}. However, computational complexity grows significantly when generating demonstrations in complex environment with a large number of agents using centralized planners. Other works consider using single-agent path planner as guidance. For instance, MAPPER~\cite{liu2020mapper} and Globally Guided RL (G2RL)~\cite{wang2020mobile} use A* for single-agent path generation.  In addition, they all apply an off-route penalty if agents failed to follow the path. However, as single-agent shortest path is usually not unique and is not global optimal for multi-agent cases, off-route penalty may mislead the agents. Another promising direction is to achieve cooperation between agents via communication. Differentiable inter-agent learning (DIAL)~\cite{foerster2016learning} uses parameter and gradient sharing for communication. Recently, graph convolution has been widely adopted to deduce the mutual interplay between agents, such as graph convolutional reinforcement learning (DGN)~\cite{jiang2018graph} and targeted multi-agent communication (TarMAC)~\cite{das2019tarmac}. 

\begin{table}[htbp]
\caption{Reward Function Design}
\label{table:reward}
\begin{center}
\resizebox{0.65\columnwidth}{!}{
\begin{tabular}{|c|c|}
\hline
\textbf{Actions} & \textbf{Reward}\\
\hline
Move (Up/Down/Left/Right)      & -0.075  \\
\hline
Stay (on goal, away goal) & 0, -0.075 \\
\hline
Collision (obstacle/agents) & -0.5 \\
\hline
Finish            & 3 \\
\hline
\end{tabular}
}
\end{center}
\end{table}

\begin{figure*}[htbp]
\centering
\begin{minipage}{0.6\textwidth}
\centering
\includegraphics[scale=0.4]{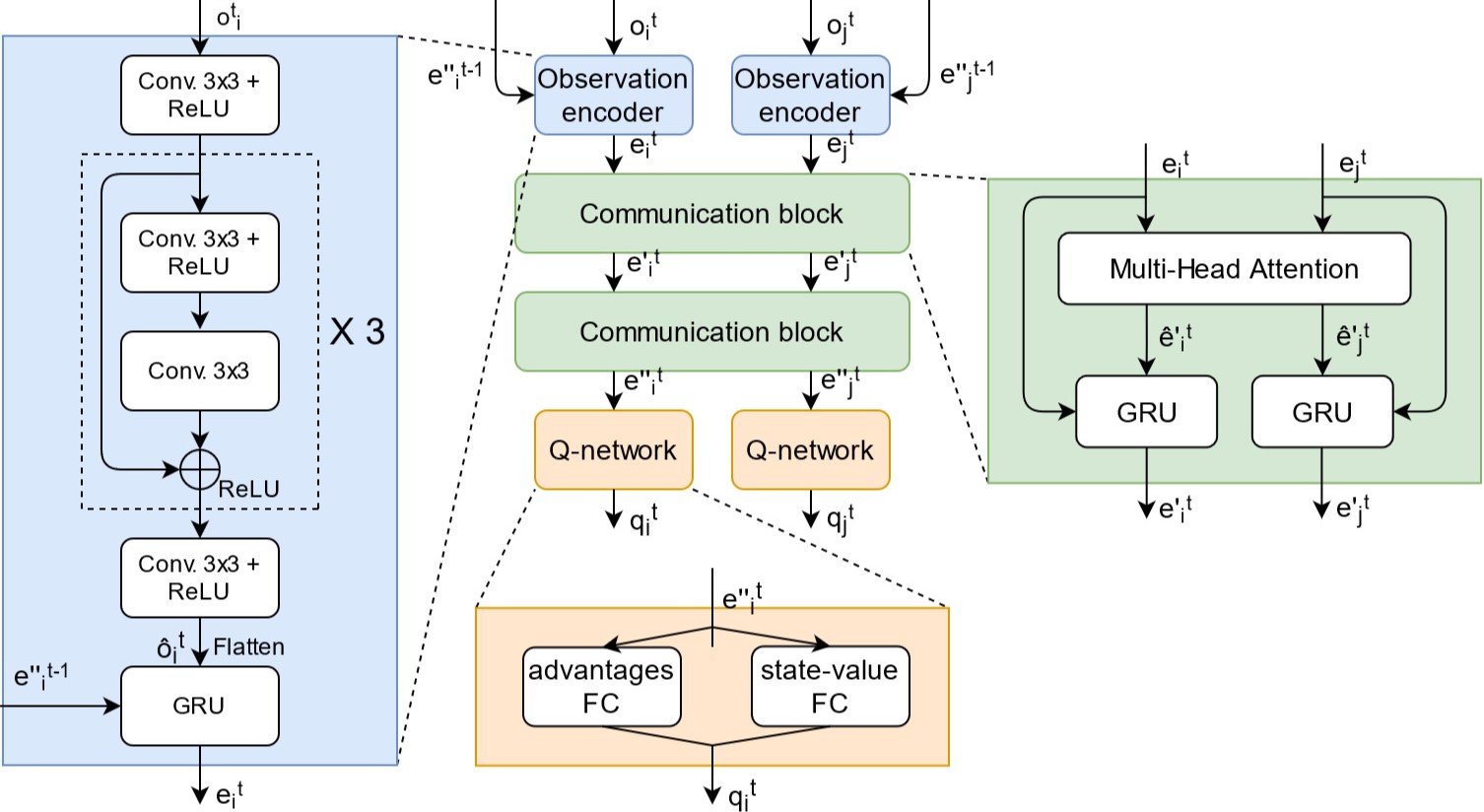}
\caption{DHC consists of three modules: observation encoder (blue), communication block (green), and Q-network (orange). The two communication blocks are identical.}
\label{fig:ach}
\end{minipage}
\begin{minipage}{0.35\textwidth}
\centering
\includegraphics[scale=0.21]{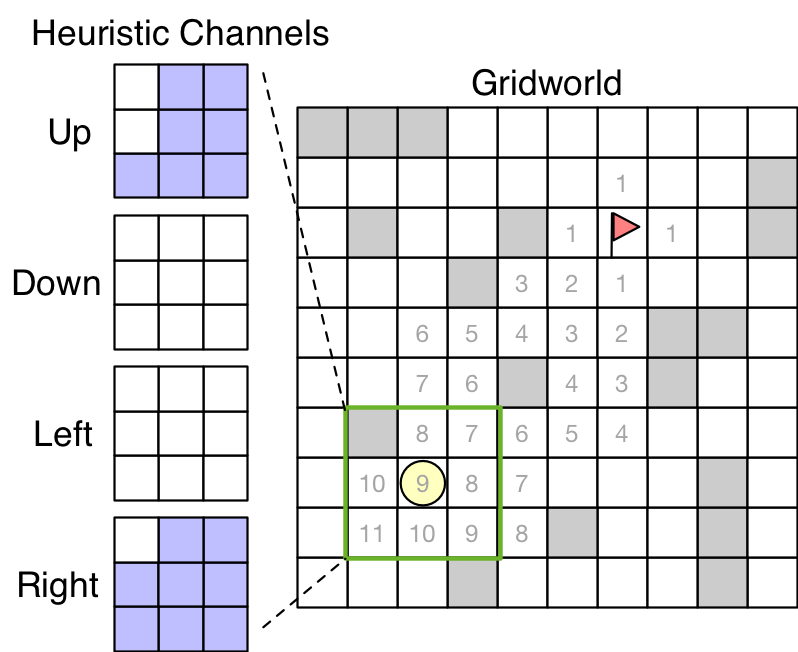}
\caption{Heuristic channels for the yellow circle agent, with $3\times 3$ FOV. The red flag is the goal, and the gray blocks are obstacles. Four channels indicate whether agent get closer to its goal by taking a certain action at those locations. Purple color means 1.}
\label{fig:nav}
\end{minipage}
\end{figure*}

\section{LEARNING ENVIRONMENT}

In this section, we introduce our environment design for MAPF. We detail the observation representation, action space and reward design of our environment. 

\subsection{Environment Setup}
We build a discrete gridworld environment for MAPF, where each agent only has partial observability. Many real world robot applications can be naturally transformed into partially observable gridworld by each agent equipped with a radar sensor to localize itself and detect surroundings. Formally, the entire space is a $m\times m$ binary matrix, where 0 represents a free location and 1 is an obstacle. Each time, $n$ beginning positions and $n$ goal positions are randomly chosen from the free locations for $n$ agents. We make sure there is no overlap among $2n$ selected positions and each goal is reachable if staring from the corresponding beginning point. Different from PRIMAL where agents are operated sequentially, our environment moves agents simultaneously and collisions are handled properly. Also, unlike MAPPER where the problem is simplified by removing agents from the environment after reaching their goals, we retain all the agents, which is more realistic for real world problem. 

\subsection{Observation Representation}
In partially observable settings, each agent can only observe the environment inside its field of view (FOV) with size $\ell\times \ell$ ($\ell<m$). We use an odd number for $\ell$ to make sure agents are at the center of FOV. The observation information is grouped into two channels. Specifically, the first channel is a binary matrix representing the obstacles inside the FOV, and the second channel is a binary matrix indicating the locations of other agents if within the FOV. We also add four heuristic channels to the input of our model (see~\ref{sec:net-design}). As the goal location of each agent can be inferred from these four channels, it is not included in the input.

\subsection{Action Space}
Agents take discrete actions in the gridworld. At each time step, agents can choose to move to the adjacent grid or stay still. We do not consider diagonal movement, thus the action space for all agents are 5. During training or execution, agents may choose invalid actions, which leads to hitting obstacles or collision between agents. PRIMAL only samples valid actions for execution, and an additional loss function is defined to force this selection. We do not filter out invalid actions. If invalid actions are taken, we recursively recover the related agents to previous states until no collision exits. Our environment design makes learning more robust compared with PRIMAL.

\subsection{Reward Design}
Motivated by the common reward design that agents are punished every step for not staying on goal to facilitate goal reaching, we design our reward function as shown in Table~\ref{table:reward}. Different from PRIMAL and MAPPER, where agents are penalized more for staying still, we treat every movement and staying (if not on the goal) as the same because in complex cases, one agent should stop and let another agent pass first in order to avoid collision.

\section{DISTRIBUTED HEURISTIC LEARNING WITH COMMUNICATION}
We present the detailed design of our model in this section. We refer to it as DHC \footnote{Code available at https://github.com/ZiyuanMa/DHC}, which represents three key components: \textbf{D}istributed, \textbf{H}euristic and \textbf{C}ommunication. The main ideas of DHC are to guide RL with heuristic and to achieve cooperation between agents by independent learning and graph convolutional communication. The independent learning schema can be naturally distributed to speed up learning procedure and achieve better performance.

\subsection{Agent Q-Network Design}\label{sec:net-design}
We build a Dueling DQN~\cite{wang2016dueling} with recurrent units to approximate agent's policy in partially observable gridworld, which maps the current observation and communication messages to Q-values of each action. The agent network consists of the following three modules: observation encoder, communication block, and Q-network, as shown in Fig.~\ref{fig:ach}. The \textit{observation encoder} consists of eight convolutional layers and a GRU.  The convolutional layers are organized as three residual blocks (each contains two convolutional layers) and two independent ones. For network of agent $i$, the local observation $o^t_i$ is first encoded into $\hat{o}^t_i$ by eight convolutional layers. Then the GRU gets $\hat{o}^t_i$ and the hidden state $e^{''t-1}_{i}$, and generates the intermediate message $e_i^t$, where $e^{''t-1}_{i}$ is the last step communication outcome. $e^t_i$ is then used for current communication between agents. The \textit{communication block} is a graph convolutional network followed by a GRU. Messages $e^t_i$ and $e^t_{\mathbb{B}_i}$ are first integrated by graph convolution, where $\mathbb{B}_i$ is the neighbors of agent $i$, to a feature vector $\hat{e}_i^t$. Then $\hat{e}_i^t$ is sent to the GRU along with $e_i^t$ serving as the hidden state to generate information $e_i^{'t}$. By communicating for multiple rounds, an agent gradually gathers more information from neighbors, which, in turn, increases the cooperation scope. We use two rounds in our model. The output from the last round works as the input for next round using the same block. The final information $e_i^{''t}$ is adopted by \textit{Q-network} to calculate Q-values for each action in a dueling manner. The state value and action advantages are separated into two streams, and then merged for the final Q function. The optimal action is chosen by the maximum Q-value. 

\subsubsection{Heuristic}Motivated by assisting RL with  guidance~\cite{sartoretti2019primal,liu2020mapper,riviere2020glas}, we introduce four heuristic channels. To reduce computational complexity, we use information provided by single-agent path planners instead of a centralized planner. Unlike MAPPER and G2RL where a specific single-agent shortest path is embedded and an off-route penalty is applied, we extract information from single source shortest paths (goal is the source) without special reward design. As the single-agent shortest path is usually not unique, we provide all the potential choices and agents are supposed to learn rational knowledge by themselves. The potential path information is embedded as follows. The four channels correspond to four actions Up, Down, Left, and Right. Each channel has the same size as the FOV, where a location is marked as 1 if and only if the agent gets closer to its goal by taking the associated action of this channel. A small case example is shown in Fig.~\ref{fig:nav}. The right action is optimal for all the locations (apart from obstacle) in FOV, thus these locations in the Right channel is marked as 1 in the figure. These four channels along with the two observation channels serve as the input of our model.

\begin{figure}[tpb]
\centering
\includegraphics[scale=0.34]{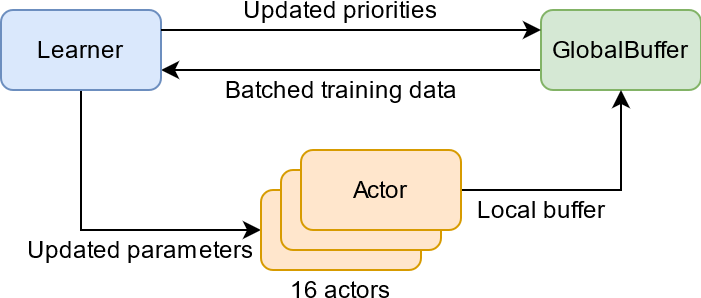}
\caption{The overview of distributed IQL framework. 16 actors generate multi-agent experiences in their own instances of environment, adding to a prioritized global buffer. The single learner updates the network and the priority of experiences.}
\label{fig:apex}
\end{figure}

\subsection{Graph Convolutional Communication}
Communication is one of the effective ways for cooperation in multi-agent systems. Motivated by the recent success of graph convolution for multi-agent communication, in this work, we treat each agent as a node, and a graph is formalized by connecting neighboring nodes. Two nodes are regarded as neighbors if they are inside the FOV of each other. Then graph convolution can be applied to derive communication. Inspired by DGN and TarMAC, we adopt multi-head dot-product attention as the convolutional kernel to compute interactions among agents. For each agent $i$, the intermediate message $e^t_i$ is projected to Query, Key and Value by matrix $\boldsymbol{W}^h_Q$, $\boldsymbol{W}^h_K$, and $\boldsymbol{W}^h_V$ in each independent attention head $h$. Let $\mathbb{B}_{i+}$ denotes the set $\{\mathbb{B}_i,i\}$, the relation between $i$ and $j\in\mathbb{B}_{i+}$ in the $h$-th attention head is computed as 

$$
\mu^h_{ij}=\mathrm{softmax}(\frac{\boldsymbol{W}^h_Q e^t_i\cdot (\boldsymbol{W}^h_K e^t_j)^{\top} }{\sqrt{d_k}}), \eqno{(1)}
$$

\noindent where $d_k$ is the dimension of Keys and $\sqrt{d_k}$ serves as a scaling factor to stabilize training. The output of each attention head for agent $i$ is the weighted summation of Values over $\mathbb{B}_{i+}$. Then the outputs are concatenated for all $\mathcal{H}$ attention heads and passed though one neural network layer $f_o$ to produce the final output of graph convolution 
$$
\hat{e}^t_i=f_o\Big(\mathrm{concatenate}(\sum_{j\in\mathbb{B}_{i+}} \mu^h_{ij}\boldsymbol{W}^h_Ve^t_j, \forall h\in\mathcal{H})\Big). \eqno{(2)}
$$

For multi-round communication (two rounds in our case), the outcome message $\hat{e}^t_i$ and the initial message $e^t_i$ are first aggregated by a GRU, then the output $e^{'t}_{i}$ acts as the initial message for the next round by repeating Equations 1-2.

\subsection{Multi-Agent Distributed Prioritized Experience Replay}
In MAPF, agents has their individual goals instead of a common goal, thus IQL is more suitable for this problem compared with centralized MARL. The appealing merit of IQL is that it avoids the scalability problem in centralized training, which requires to learn a Q-function for joint actions over all agents. The joint action space grows exponentially as the number of agents increases. On the other hand, IQL is naturally appropriate to learn decentralized policy for partially observable settings, because each agent makes decision only based on its own observation. As each agent in the MAPF environment plays the same role as others, to simplify the training process, instead of learning multiple policies for multiple agents, we train a single model from a single agent's perspective while treating others as part of its environment. The final trained policy can be applied to each agent for decentralized execution.

To speed up deep RL with the advancement of computational resources, many RL algorithms have been promoted by distributed training, such as A3C~\cite{mnih2016asynchronous} and Distributed PPO~\cite{heess2017emergence}. The idea behind these distributed versions of algorithms is parallelizing the gradient computation so as to facilitate parameter optimization. Specifically, PRIMAL is distributed using A3C as a backend. Another direction to distributed RL is parallelizing experience data generation and selection with a shared replay memory, such that more high priority data can be gathered to benefit the model. This is referred to as Ape-X architecture, where multiple actors generate experience and a single learner updates the network~\cite{horgan2018distributed}. Our IQL from a single agent's perspective method can be naturally distributed by Ape-X framework. 

Fig.~\ref{fig:apex} illustrates our system flow. In the experiments, we setup sixteen independent actors running on CPUs to generate data, and a single learner on GPU to train. Each actor has a copy of the environment with current Q-network and keep generating new transitions from multiple agents and initializing priorities for them. The transitions from all actors are fed into a shared prioritized replay buffer. Then learner samples the most useful experiences from the buffer and updates the network and priorities of the experience. Note that although the model is trained for a single agent, the transitions of all the agents need to be stored for communication purpose, and the priorities are initialized and updated from that agent's perspective. As priorities are shared, the good experiences explored by any actor can improve the learner. 

The final loss function is a multi-step TD error
$$
\mathcal{L}(\theta)= \mathrm{Huber}(R_t - Q(s_t,a_t,\theta)) \eqno{(3)}
$$
with $R_t = r_{t}+\gamma r_{t+1}+\ldots +\gamma^n Q(s_{t+n},a_{t+n},\bar{\theta})$, where $R_t$ is the total return of the agent we care about, $s_t$ and $a_t$ are the state and action of that agent, and $\bar{\theta}$ denotes the parameters of the target network, a periodical copy of the online parameters $\theta$.

\section{EXPERIMENTS}
Learning directly from a large size environment with lots of agents is hard. Instead, we use curriculum learning method by gradually introducing more difficult tasks to agents~\cite{bengio2009curriculum}. Starting from a easy task with only one agent in a $10\times 10$ environment, we establish two new challenging tasks for agents by increasing the agent amount by one or increasing the environment size by five, if the success rate of the current task exceeds 0.9. As the training scale grows, the final task with twelve agents in a $40\times 40$ environment is reached.

In real world problem, the communication latency are restrictive and bandwidth is limited. Full communication with neighbors leads to huge communication overhead and latency. In the experiment, we choose the nearest two neighbors for communication. In addition, to make the policy more robust, the agent that we are focusing on takes $\epsilon-$greedy actions during training, while no random action is taken by other agents to make the environment more stationary.

For other parameter settings during training, the obstacle density of the environment is sampled from a triangular distribution between 0 and 0.5 with a peak at 0.33 (same as PRIMAL). The FOV size is $9\times 9$ ($10\times 10$ in PRIMAL, we make it odd). The maximum episode length is 256. We train the network with a batch size of 192 and a sequence length of 20 (limit by memory). We use a dynamic learning rate beginning at $10^{-4}$ and decreasing by fifty percent at 100k step and 300k step. The maximum training step is 500k.

\subsection{Success Rate and Average Step}
We examine the performance of our model and one of the state-of-the-art methods, PRIMAL, with respect to success rate and average step. Success rate measures the ability to complete a task within the given time steps. Average step measures the average time consumed for a task, where smaller value indicates better policy. (We average all cases to calculate average step). PRIMAL uses a very similar environment as ours, making the results comparable. MAPPER does not stick to standard MAPF setting, so we do not compare with it here. We set up two different maps, $40\times 40$ and $80\times 80$ for testing. To highlight the ability of DHC to handle obstacle rich environments, we set the obstacle density $=0.3$, the highest density used in PRIMAL. For each agent number $\{4, 8, 16, 32, 64\}$, we set up 200 test cases. The maximum time step for $40\times 40$ map is $256$, and $386$ for $80\times 80$ map, the same as the PRIMAL's setting. We also include ODrM*~\cite{wagner2015subdimensional}, a centralized planner, as a reference for average step.

\begin{figure}[tp]
\begin{minipage}{.5\columnwidth}
\centering\
\subfloat{\label{main:a}\includegraphics[scale=.205]{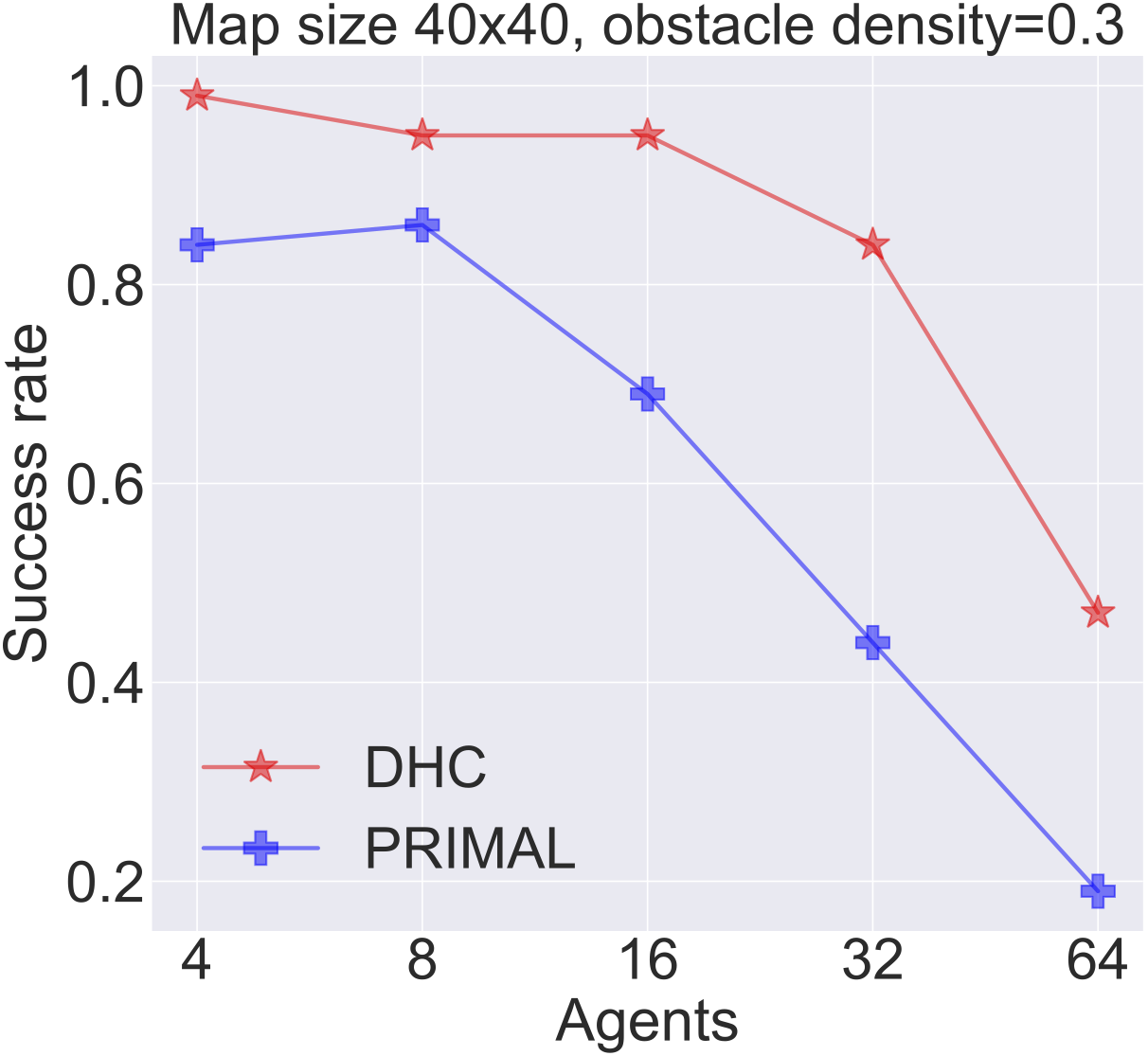}}
\end{minipage}%
\begin{minipage}{.5\columnwidth}
\centering
\subfloat{\label{main:b}\includegraphics[scale=.205]{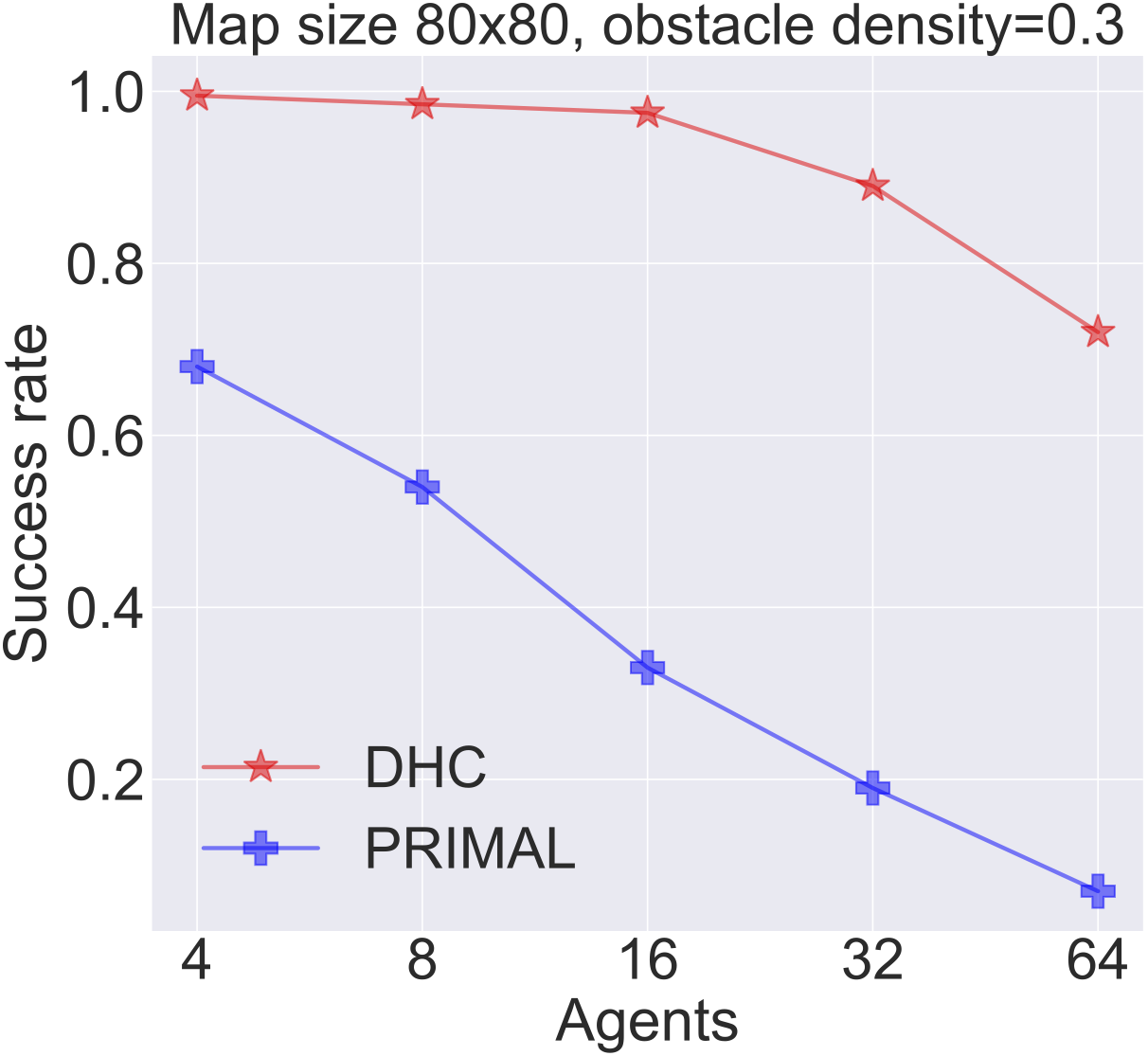}}
\end{minipage}
\caption{Success rate of our method compared with PRIMAL in two different scenarios.
}
\label{fig:vs_primal}
\end{figure}

Fig.~\ref{fig:vs_primal} shows the success rate of our method compared with PRIMAL in these two different environments. Our method outperforms PRIMAL in all the cases. PRIMAL preforms better in relatively small environment ($40\times 40$) than larger one ($80\times 80$), which shows the IL part of PRIMAL does not deliver a good guidance to its RL component during learning, thus it suffers performance degradation on long-horizon task. By self-learning potential shortest paths from the heuristic guidance, our model can find paths for agents in larger map much easier. In congested cases where a large number of agents operate in a small environment, for example, $32$ and $64$ agents in $40\times 40$ map, PRIMAL can not handle collisions properly. In contrast, DHC teaches agents to cooperate with neighbors via communication, which leads to coordinated behaviours and higher success rates. Table~\ref{table:avg_step} verifies these two merits of DHC with respect to the average step. DHC requires a less step increment to find paths when switches from $40\times 40$ map to $80\times 80$ map, which indicates DHC learns useful information from heuristic to better identify paths in larger environment. In addition, DHC always consumes less time steps to finish tasks, showing that the communication helps to handle collisions and accelerates path finding.

\begin{table}[tbp]
\caption{Average Step in two different environments with obstacle density $=0.3$}
\label{table:avg_step}
\begin{center}
\resizebox{0.97\columnwidth}{!}{
\begin{tabular}{c|c|cc|c|cc}

\hline
Average Step &\multicolumn{3}{|c}{Map size $40\times 40$}      & \multicolumn{3}{|c}{Map size $80\times 80$}  \\
\hline
 Agents &  ODrM* & DHC & PRIMAL & ODrM* & DHC & PRIMAL\\
\hline
    4   & 50.00&\textbf{52.33}                   &79.08      & 93.40 &  \textbf{96.72}                  &     134.86\\

    8   & 52.17&\textbf{63.90} & 76.53           & 104.92 &\textbf{109.24} & 153.20\\

   16   & 59.78&\textbf{79.63} & 107.14                     & 114.75&\textbf{122.54} & 180.74\\

   32   & 67.39&\textbf{100.10} & 155.21         & 121.31 &\textbf{138.32} & 250.07\\

   64   & 82.60&\textbf{147.26} & 170.48         & 134.42 &\textbf{163.50} & 321.63\\
\hline
\end{tabular}
}
\end{center}
\end{table}

\subsection{With and Without Heuristic or Communication}

To further assess the functionality of the heuristic and communication components of our method, we develop a baseline model and a DHC variant. The baseline model does not contain heuristic channels or communication block. However, we do not simply remove these two parts from DHC to form the baseline, instead, to provide sufficient information of the environment and goal locations to agents, we modify the input to have the same format as PRIMAL, which contains the obstacle positions, the nearby agent positions, the projected goal positions of nearby agents (project to the boundary of FOV if outside of FOV), and the position of the agent's own goal is within the FOV. The DHC variant, denoted by DHC/Comm, uses the same input as DHC but omits the communication block (graph convolution). These two models are both distributed during training and use the same training parameters as DHC.

\begin{figure}[tp]
\begin{minipage}{.5\columnwidth}
\centering\
\subfloat{\label{main:a}\includegraphics[scale=.205]{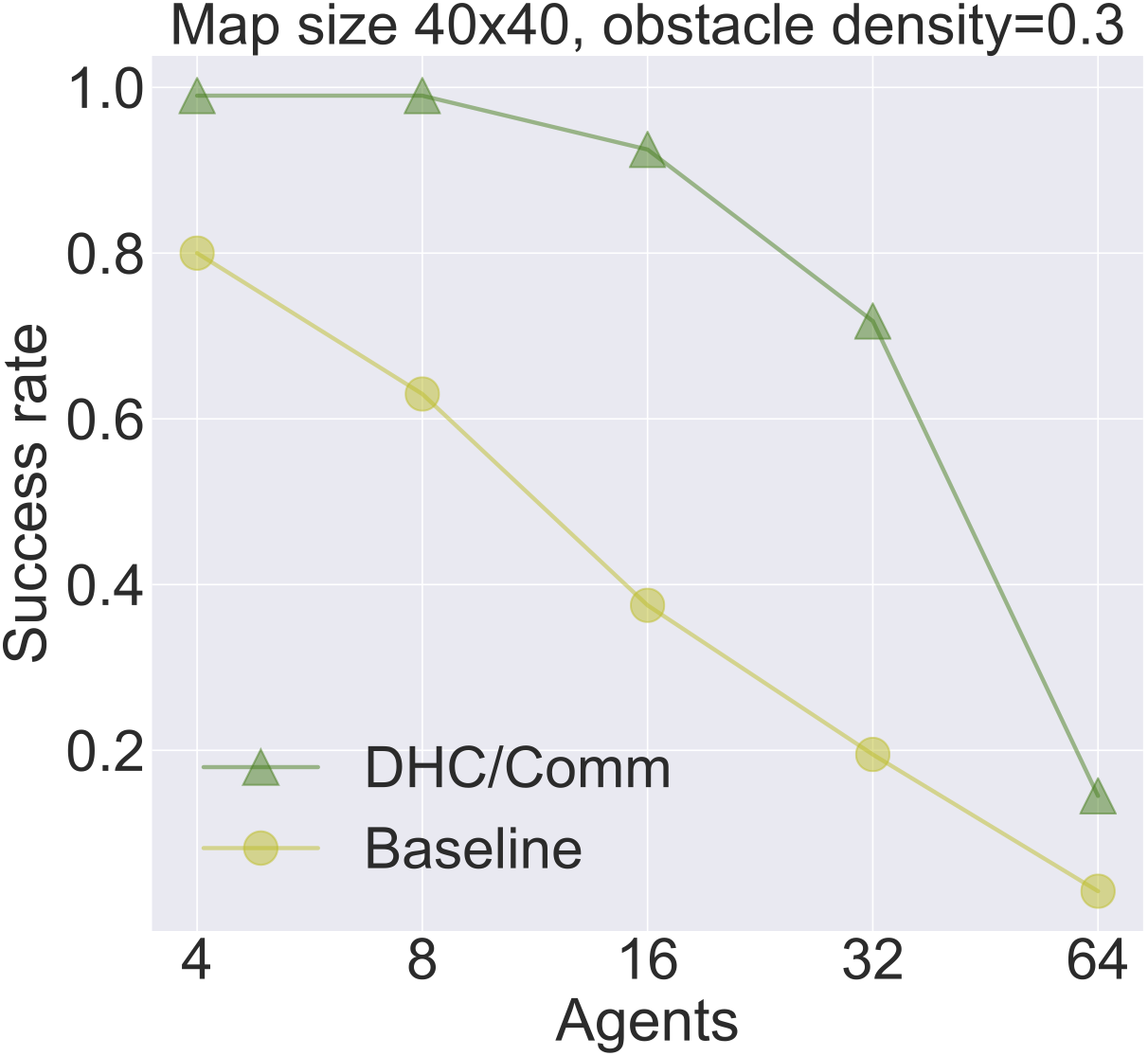}}
\end{minipage}%
\begin{minipage}{.5\columnwidth}
\centering
\subfloat{\label{main:b}\includegraphics[scale=.205]{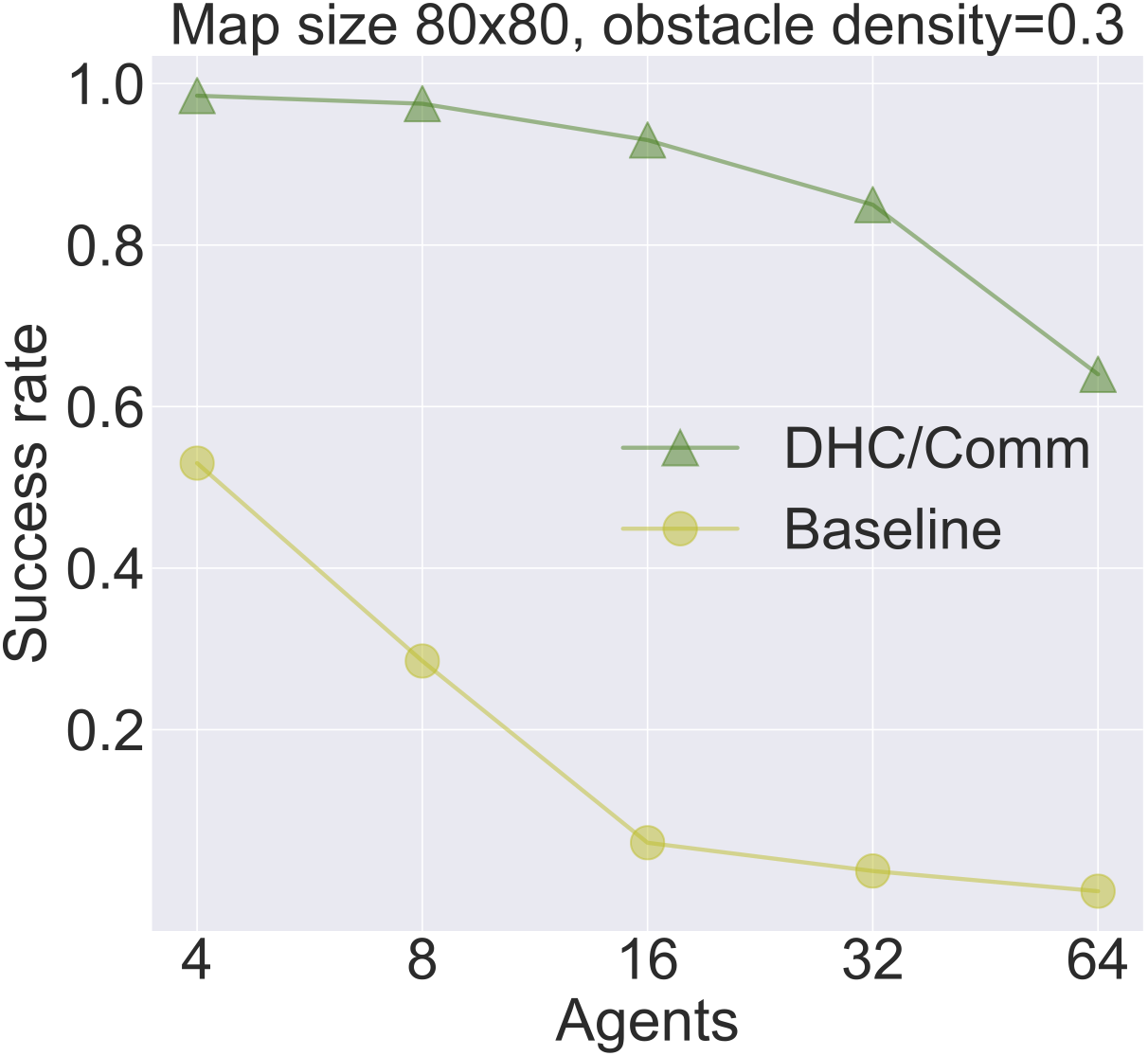}}
\end{minipage}\par\smallskip
\begin{minipage}{.5\columnwidth}
\centering
\subfloat{\label{main:c}\includegraphics[scale=.205]{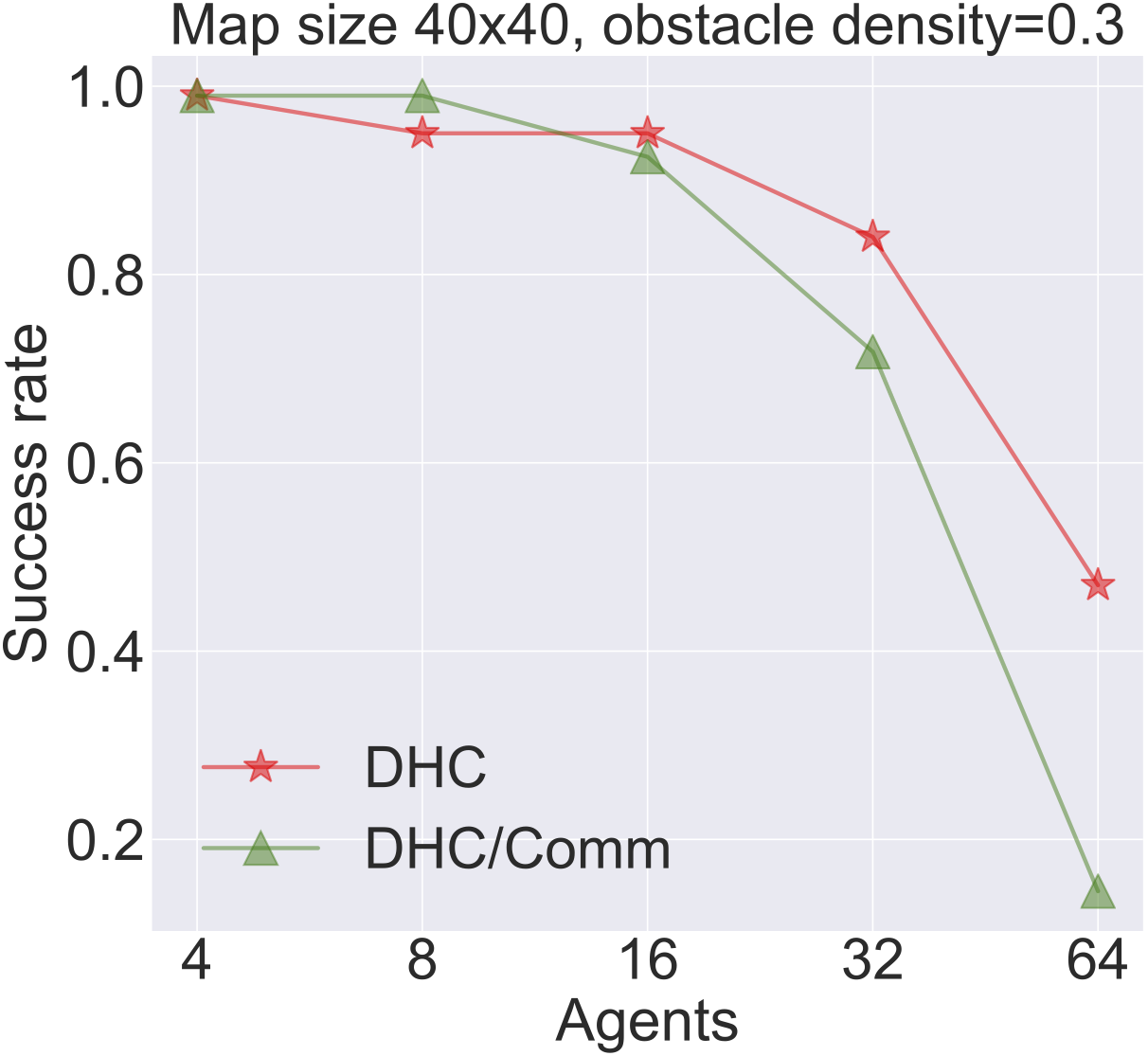}}
\end{minipage}%
\begin{minipage}{.5\columnwidth}
\centering
\subfloat{\label{main:d}\includegraphics[scale=.205]{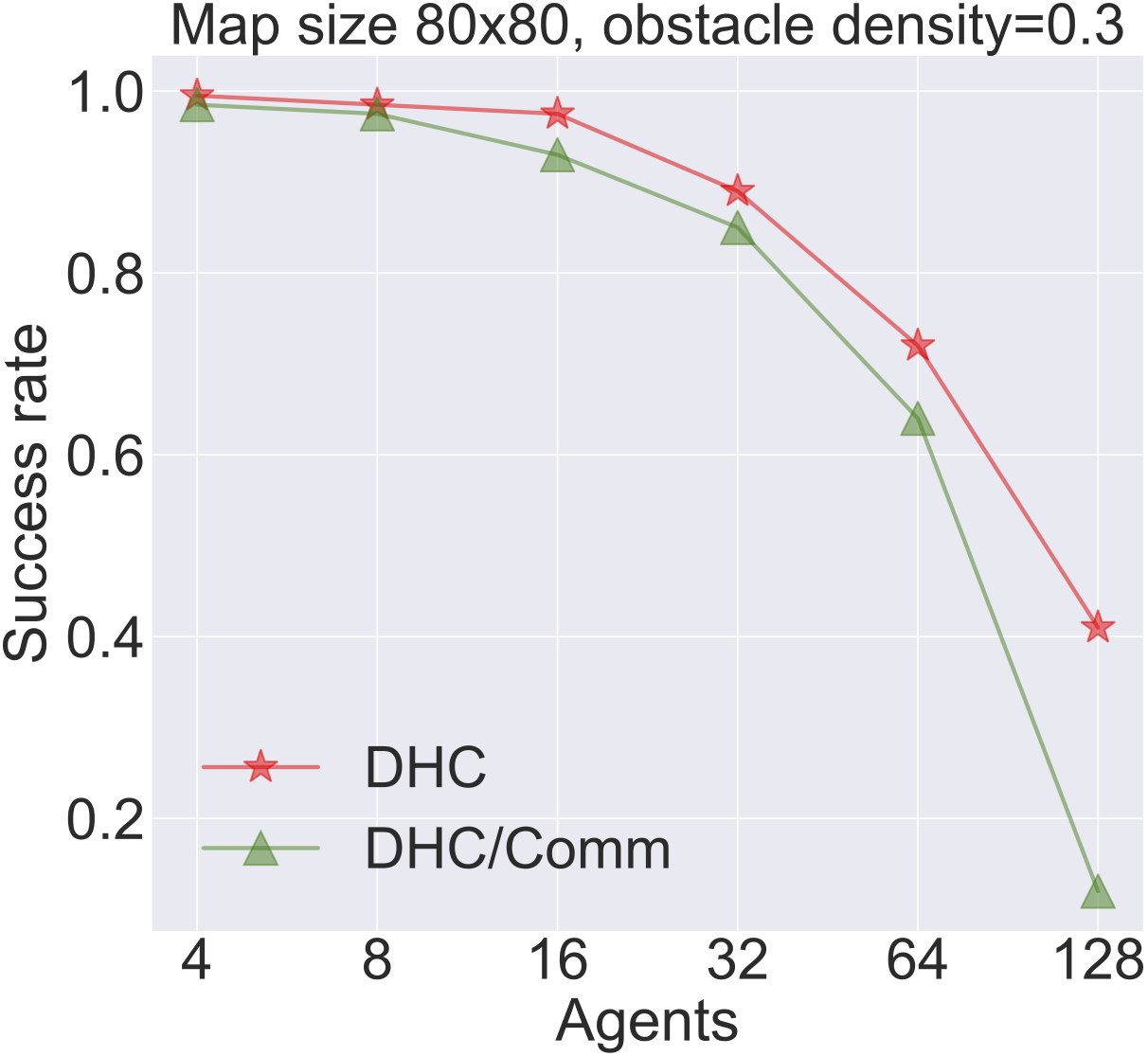}}
\end{minipage}
\caption{Success rates of baseline, DHC/Comm, and DHC with respect to heuristic channels and communication block.
}
\label{fig:heur-comm}
\end{figure}

We compare the success rate of DHC/Comm with baseline model as shown in the first row of Fig.~\ref{fig:heur-comm} to demonstrate the reasonable design of the heuristic channels. Although the baseline model uses information rich input, the lack of learning rational knowledge from heuristic guidance leads to a sharp decrease of the performance. Also, the absence of heuristic makes baseline model behave worse in long-horizon tasks in larger environment. To reveal the capacity of communication, we compares DHC with DHC/Comm as shown in the second row of Fig.~\ref{fig:heur-comm}. When the agent density is low, for instance, $4$ and $8$ agents in both maps, there is little difference between their performances, because agents have less chances to encounter with each other and thus less cooperation is required. In tight environments, such as $64$ agents in $40\times 40$ map and $128$ agents in $80\times 80$ map, DHC achieves much higher success rates due to the communication ability.


\section{CONCLUSIONS}

We propose a distributed heuristic deep Q-learning method with communication for MAPF with standard MAPF environment settings. Our aims are to deliver guidance to RL by self learning so as to handle long-horizon goal-oriented tasks, and to achieve cooperation between agents through communication for coordinated behaviours in congested situations. We introduce the embedding of shortest paths from single source instead of a specific single-agent shortest path as heuristic to incorporate guidance in RL. Graph convolution mechanism is adopted as communication to gain cooperation between agents. The model is distributed and trained from a single agent's perspective, leading to good scalability. In experiments, our method is more adequate for long-horizon tasks and achieves highest success rate with lowest average step in both sparse and tight environments compared with PRIMAL. As we fix the number of communication agents, an important direction for future work is to increase communication scope while reduce the  overhead and latency.






\section*{ACKNOWLEDGMENT}

This work was supported by Natural Sciences and Engineering Research Council under grant RGPIN-2020-06540.


\bibliographystyle{./IEEEtran} 
\bibliography{./IEEEabrv,./IEEEexample}

\end{document}